\documentclass[letterpaper, 10 pt, conference]{ieeeconf}  

\IEEEoverridecommandlockouts                              

\usepackage{graphics} 
\usepackage{epsfig} 
\usepackage{mathptmx} 
\usepackage{times} 
\usepackage{amsmath} 
\usepackage{amssymb}  

\usepackage{multirow}

\def\etal{\emph{et al.}}

\newcommand{\argmin}{\mathop{\rm argmin}\limits}

\renewcommand{\tilde}{\widetilde}

\def\sectionskip{\vspace{0pt}}
\def\subsectionskip{\vspace{0pt}}
\def\figureskip{\vspace{-4pt}}
\def\tableskip{\vspace{0pt}}

\title{\LARGE \bf
Virtual Inverse Perspective Mapping for \\ Simultaneous Pose and Motion Estimation
}

\author{Masahiro Hirano$^{1}$, Taku Senoo$^{2}$, Norimasa Kishi$^{1}$ and Masatoshi Ishikawa$^{3}$
  \thanks{$^{1}$Masahiro Hirano and Norimasa Kishi are with Institute of Industrial Science, The University of Tokyo, Tokyo 153-8505, Japan.
  {\tt\small mhirano@iis.u-tokyo.ac.jp}}%
  \thanks{$^{2}$Taku Senoo is with Faculty of Information Science and Technology,
Hokkaido University, Hokkaido 060-0814, Japan}
  \thanks{$^{3}$Masatoshi Ishikawa is with Information Technology Center, The University of Tokyo, Tokyo 113-8656, Japan.}
}

\begin{document}

\maketitle
\thispagestyle{empty}
\pagestyle{empty}

\begin{abstract}
We propose an automatic method for pose and motion estimation against a ground surface for a ground-moving robot-mounted monocular camera.
The framework adopts a semi-dense approach that benefits from both a feature-based method and an image-registration-based method by setting multiple patches in the image for displacement computation through a highly accurate image-registration technique.
To improve accuracy, we introduce virtual inverse perspective mapping (IPM) in the refinement step to eliminate the perspective effect on image registration.
The pose and motion are jointly and robustly estimated by a formulation of geometric bundle adjustment via virtual IPM.
Unlike conventional visual odometry methods, the proposed method is free from cumulative error because it directly estimates pose and motion against the ground by taking advantage of a camera configuration mounted on a ground-moving robot where the camera's vertical motion is ignorable compared to its height within the frame interval and the nearby ground surface is approximately flat.
We conducted experiments in which the relative mean error of the pitch and roll angles was approximately $1.0$ degrees and the absolute mean error of the travel distance was $0.3$ mm, even under camera shaking within a short period.
\end{abstract}

\section{Introduction}
\sectionskip
Recognition of the surrounding environment is one of the most important features in developing ground-moving robots such as automobiles.
Among various sensors, cameras have been widely applied owing to their versatility.
For instance, lane departure warning systems are an increasingly standard feature in modern vehicles; these systems recognize a traffic lane via cameras and warn the driver when the vehicle deviates from the lane. To recognize the relative positional relation between the camera and lane, the camera pose with respect to the ground must be known. 
Many other applications, including mobile robot navigation~\cite{Lee16}, obstacle detection~\cite{Bertozzi98a}, and visual odometry~\cite{Lovegrove11}, rely on the camera pose against the ground.
For these applications, the estimation accuracy of the camera pose directly influences recognition performance. In many cases, the pose is estimated in advance through a carefully engineered calibration technique; however, this approach is insufficient because the pose can dynamically change owing to braking and acceleration maneuvers and road-surface irregularities, as discussed in \cite{Westerhoff16}.

Inertial measurement units are commonly used to provide an external reference for pose estimation, but
they tend to suffer from accumulated drift; thus, their estimation accuracy deteriorates over a period of extended operation. 
Therefore, estimation of the relative pose with respect to the ground through only a captured image sequence is desired.
Once an automatic estimation system is implemented, the camera system can be automatically recalibrated if it suffers from rapid shaking or gradual change over time.
In this study, we addressed an {\it autocalibration} problem: automatic estimation of a monocular camera pose from captured image sequences.
We focused on estimating extrinsic parameters with respect to the ground, which can dynamically change during camera motion.
We further estimated the camera motion simultaneously by making maximum use of the captured images.

Recent advances in camera pose estimation in robotics have been made in the field of {\it visual odometry} (e.g.~\cite{Nister04, Forster17}), which jointly estimate camera pose and motion with respect to the initial pose of the camera, not the ground.
Pose against the ground could be estimated by assigning premeasured offsets by means of standard calibration but it inevitably suffers from cumulative error.
The proposed method, in contrast, directly estimates camera poses with respect to the ground at every moment and achieves highly accurate, cumulative error-free estimation by taking advantage of a ground-moving robot-mounted camera configuration; the camera constantly observes the nearby ground, which can be regarded as approximately flat.
Under this assumption, the ground surface texture becomes a great reference for estimation.
Additionally, the camera's vertical motion within a frame interval is assumed to be much smaller than the height of the camera; the camera height within a frame interval can be approximated as constant.

In this work, the proposed method benefits from two approaches: by adopting a semi-dense approach, a set of patches were chosen to compute their apparent motion via an image-registration technique, and then optimization was performed by means of robust geometric bundle adjustment to obtain pose and motion estimations.
We employed a rigid image-registration method, phase-only correlation (POC) \cite{Chen94, Takita03}, which is known for its high accuracy and robustness.
Such rigid image-registration performance is maximized when the patches undergo rigid transformations despite the transformations between consecutive patches usually subject to the perspective effect. Therefore, we propose applying inverse perspective mapping (IPM) \cite{Bertozzi98b} according to the initial estimated pose to generate roughly perspective-free patches, which we call {\it virtual IPM}.
We formulated a dedicated geometric bundle adjustment problem for virtual IPM based on POC-estimated translations.

We carefully designed ground-truth and comparative evaluation experiments and demonstrated that camera pose and motion estimation was improved through virtual IPM and achieved very high accuracy with only two consecutive frames, without prior pose and motion, even under dynamic shaking.

\sectionskip
\section{Related work}
\sectionskip
Because pose and motion estimation for ground-moving robots has been vastly investigated, this section aims to provide a brief overview of previous works that leverage the global/local planarity of the ground. Online estimation of camera pose against the ground has drawn attention in the area of moving robots, including automobiles. Miksch \etal \cite{Miksch10} and Knorr \etal \cite{Knorr13} devised a method to extract feature points from the ground and robustly estimate plane-induced homography between consecutive frames under the assumption that the road is a flat surface. The pose was computed by decomposing the homography matrix. These methods are robust against initial pose estimation but rely on feature extraction that could deteriorate in textureless-dominant surfaces.
Other works used image-registration techniques to extract the apparent motion of an image caused by camera movement. Westerhoff \etal \cite{Westerhoff16} and Zienkiewicz and Davison \cite{Zienkiewicz15} estimated image-warping parameters between consecutive frames by iteratively minimizing the photometric error. These methods are fast and lightweight but need good initial pose estimation to converge to the optimum. They potentially suffer from a strong road marking that causes a problem similar to an {\it aperture problem} because they use a single image region to match warped road regions.

Nister \etal \cite{Nister04} introduced visual odometry that jointly estimates instantaneous change in pose from a captured image sequence.
Campbell \etal \cite{Campbell05} utilized optical flow to estimate ego motion with respect to the ground. Similar to methods that use feature points in autocalibration, optical flow extraction does not always succeed in obtaining textureless or high-frequency images. Nourani-Vatani \etal \cite{NouraniVatani09} and Kitt \etal \cite{Kitt11} proposed visual odometry methods based on template matching. Lovegrove \etal \cite{Lovegrove11} employed a more sophisticated image-registration technique called efficient second-order minimization \cite{Malis04} for improved estimation. Zienkiewicz and Davison \cite{Zienkiewicz15} extended their work to simultaneously execute autocalibration.

In contrast to these methods, the proposed method sets multiple image patches to compute each displacement to gain robustness against initial pose estimation and employs a phase-based, highly accurate image-registration technique aiming at improving estimation accuracy. This semi-dense approach is adopted by a state-of-the-art visual odometry method~\cite{Forster17}, against which we compare our work in the experiments.
We employ a refinement process that removes the perspective effect to maximize the performance of the image-registration technique.

\begin{figure*}[!t]
    \centering
    \includegraphics[width=\linewidth, keepaspectratio]{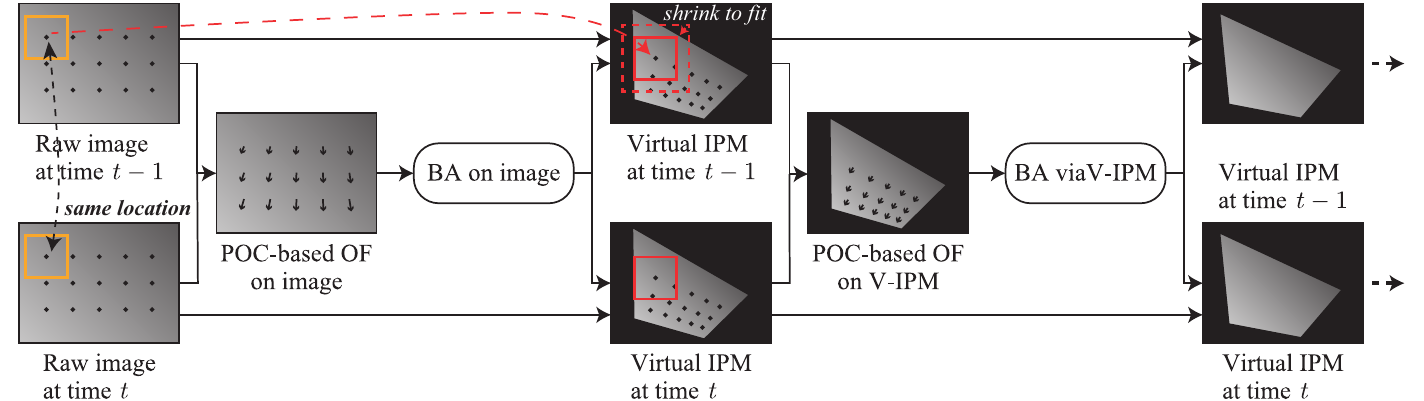}
    \caption{Workflow of the proposed method. Estimates are refined via virtual IPM. (OF: optical flow, BA: bundle adjustment, V-IPM: virtual IPM)}
    \label{fig:workflow}
    \figureskip
\end{figure*} 

\begin{figure}[t]
    \centering
    \includegraphics[width=\linewidth]{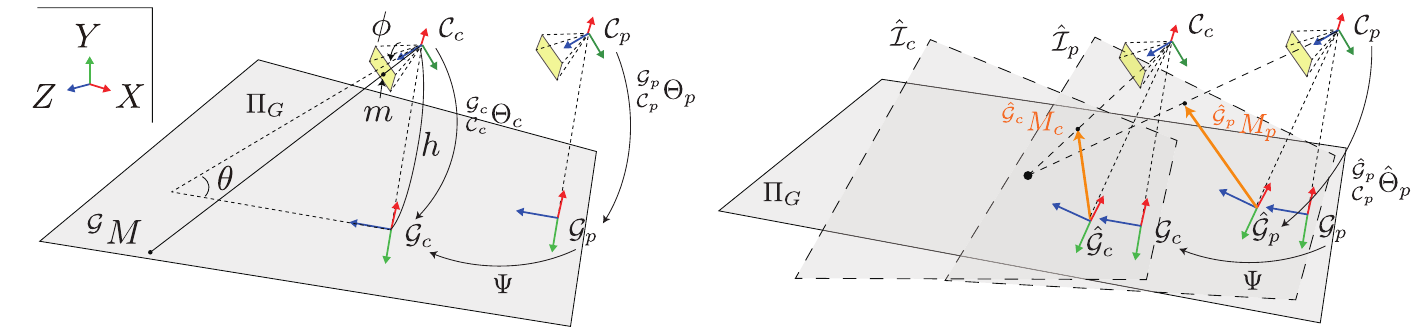}
    \caption{Parameterization.}
    \label{fig:parameterization}
    \figureskip
\end{figure} 

\sectionskip
\section{Autocalibration and motion estimation through multiple patch registration}
\sectionskip
In Fig.~\ref{fig:workflow}, the overall sketch of the proposed method is shown.
In the proposed method, we extensively used the dynamic appearance change of the ground-surface texture to robustly and accurately estimate camera pose and motion with respect to the ground.  
We first selected a set of points on the captured image, termed {\it interest points}. For each interest point, we cropped a fixed-size square patch centered at the same location (interest point) from two consecutive frames. 
We observed that the computed motion vectors differed according to their location on the image. This implied that the motion vector field encoded information about the camera pose. Furthermore, we estimated camera motion parallel to the ground from the magnitude of the motion vector field once we assumed a predetermined scale. Based on this observation, we estimated the pose and motion by formulating an optimization problem, which minimized the sum of squared residuals between a motion vector field reconstructed by the estimated pose and motion parameters and that computed via the image-registration technique.
This framework is often called {\it geometric bundle adjustment}.

These patches usually suffer from the perspective effect induced by a nonvertical camera pose, which degrades accuracy in rigid 2D-motion estimation. To ameliorate this issue, we propose to apply IPM to reduce the perspective effect, resulting in a more accurate motion-vector-field estimation. First, we set a {\it virtual IPM plane} based on current estimates, not the true pose, and then computed displacements through POC on this plane. Because the motion vector field was computed on the virtual IPM plane instead of the image plane, we reformulated the bundle adjustment to minimize reprojection errors on the virtual IPM plane. We obtained more accurate estimates via refinements, as described earlier. The proposed method is less dependent on the initial value of the optimization process and provides good estimates owing to the accurate image-registration technique, thus making maximum use of the local ground planarity.

\subsectionskip
\subsection{Parameterization}
Let $m$ be a 2D point on an image represented in the image coordinates and 
$\tilde{m}$ be its corresponding representation in homogeneous coordinates (See Fig.~\ref{fig:parameterization}).
Three-dimensional (3D) point $M$ represented in coordinate system $\mathcal{A}$ is denoted by $^\mathcal{A} M$,
and its corresponding representation in homogeneous coordinates is represented by ${}^\mathcal{A} \tilde{M}$.
A standard image-acquisition process associates ${}^{\mathcal{A}} \tilde{M}$ 
with an image point $\tilde{m}$ via perspective mapping
$\tilde{m} = \pi({}^{\mathcal{A}} \tilde{M}) = K {}_{\mathcal{A}}^{\mathcal{C}}H {}^{\mathcal{A}}\tilde{M},$
where $K$ represents a camera intrinsic matrix, and
${}_{\mathcal{A}}^{\mathcal{C}}H$ represents the transformation matrix from $\mathcal{A}$ to the camera's coordinate system $\mathcal{C}$.
Assuming that every camera pixel captures a point on the ground plane $\Pi_G$,
we can construct a bijective mapping between camera plane $\Pi_C$ and ground $\Pi_G$.
IPM can be regarded as the mapping from camera plane $\Pi_C$ to ground $\Pi_G$ induced by the projection relationship,
often called a {\it bird's-eye view}.

We define the IPM in a geometric manner.
Let $\mathcal{G}$ be a ground coordinate system, whose origin is the foot of a perpendicular from the origin of camera coordinate system $\mathcal{C}$ to ground $\Pi_G$; its $z$-axis is parallel with the projected $z$-axis of $\mathcal{C}$ to $\Pi_G$, as depicted in Fig.~\ref{fig:parameterization}.
We can characterize ground coordinate system $\mathcal{G}$ with respect to camera coordinate system $\mathcal{C}$ by using parameter set ${}_\mathcal{C}^\mathcal{G} \Theta$, which includes pitch angle $\theta$, roll angle $\phi$, and the height from the ground, $h$. 
The IPM of point $m \in \Pi_C$ to ground plane $\Pi_G$ is described as 
${}^\mathcal{G} M = \pi^{-1}(m;{}_\mathcal{C}^\mathcal{G}\Theta).$

Next, we formulate the autocalibration problem.
Fig.~\ref{fig:parameterization} illustrates the parameterization of the formulation. 
We assumed the camera's vertical motion during the short frame interval was much smaller than camera height.
We approximated the height of the camera from the ground $h$ to be constant within the frame interval.
Because we aimed to estimate relative pitch and roll angles with respect to the ground, the previous and current moment's pitch and roll angles $\Theta = \{ {}_{\mathcal{C}_p}^{\mathcal{G}_p}\Theta_p, {}_{\mathcal{C}_c}^{\mathcal{G}_c}\Theta_c\}$ were estimated with respect to ground coordinates $\mathcal{G}_p, \mathcal{G}_c$, respectively. 
The camera motion was parameterized by the two-dimensional change $\Psi = (t_x,t_z,\psi)$, where $t_x$ and $t_z$ represent translational changes in the $x$- and $z$-axes, respectively, and $\psi$ represents the change in the yaw angle. 
Because the absolute yaw angle can be parameterized by a predefined coordinate, we estimate only the change in yaw angles.
To summarize, the pose and motion are parameterized by $\{\Theta, \Psi\}$, which have seven degree of freedoms.

Note that, the relative position of two poses for a freely moving camera generally can be parameterized by six parameters;
however, in our formulation, we must add one more parameter that determines the relative positional relation between the camera and the ground because the consistency condition of camera height does not identify a common tangent plane (i.e., the ground) of both balls centered at the camera locations. 

\begin{figure*}[t]
    \centering
    \includegraphics[width=.9\linewidth]{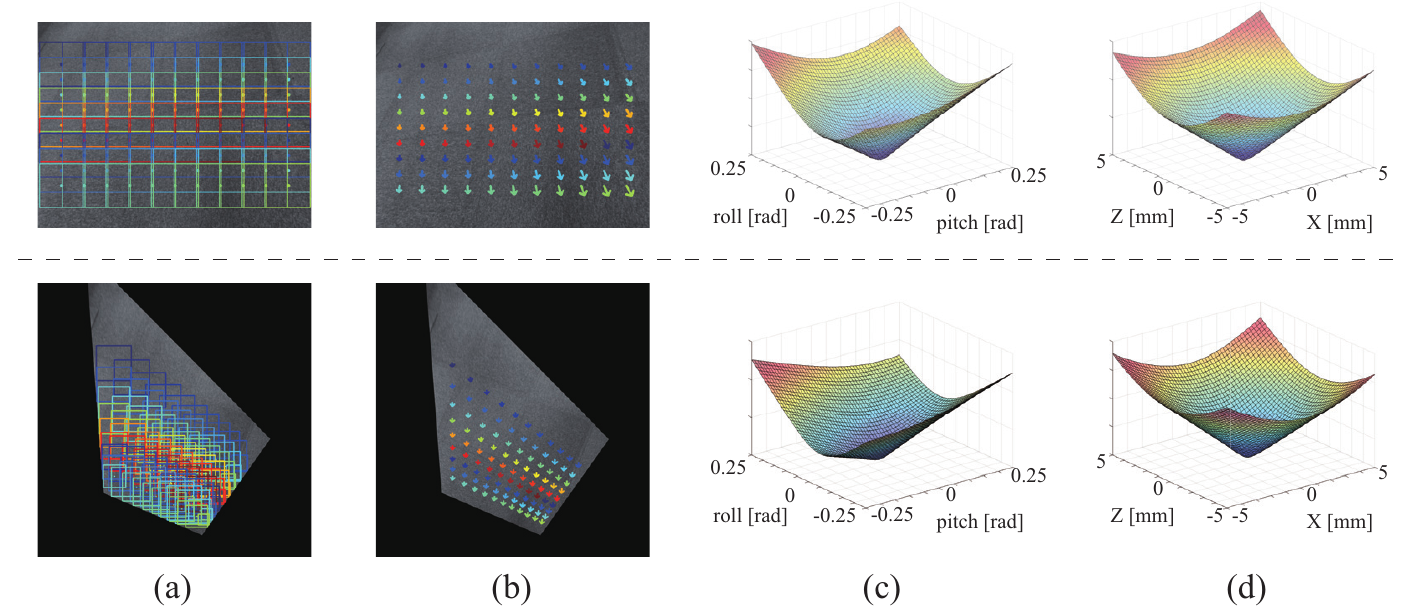}
    \figureskip
    \caption{Processes in initial estimation (upper) and refinement (lower). (a) Patches are cropped in the images, and (b) displacements are computed for each patch. Patches are illustrated in corresponding colors for easy illustration. (c) Postural and (d) translational slices of the objective function, whose axes denote the difference between the true and estimated parameters.}
    \label{fig:process}
\end{figure*} 

\subsectionskip
\subsection{Initial estimation on image plane}
We select a set of interest points $m_i$ on image plane $\mathcal{I}_p$.
This distribution can be arbitrary, so we utilized a semantic segmentation technique to select only a point from the ground region.
In this work, we chose equally distributed points on $\mathcal{I}_p$, forming a uniform grid structure, as depicted in Fig.~\ref{fig:workflow}.
The number of interest points was also selected depending on the required application accuracy.
For each interest point, a region of interest (ROI) for displacement computation was selected, whose dimension was determined such that the whole region of the ROI lay inside the image.
For previous image $\mathcal{I}_p$ and current image $\mathcal{I}_c$, we cropped a pair of patches corresponding to the same interest point and computed the apparent motion of the point through POC.
A motion vector field was generated by gathering all pairs of patch displacements, as shown in Fig.~\ref{fig:workflow}.

We estimated the initial camera pose and motion by exploiting the motion vector field by minimizing reprojection residuals.
Let $m_p$ be an interest point on previous frame $\mathcal{I}_p$ and $d$ be its corresponding displacement vector computed by POC.
Point $m_p$ is inversely projected to a 3D point on the ground 
${}^{\mathcal{G}_p}M = \pi^{-1}(m_p;{}_{\mathcal{C}_p}^{\mathcal{G}_p}\Theta_p)$.
This point can be represented in the current frame's ground coordinate as 
${}^{\mathcal{G}_c}M = {}_{\mathcal{G}_p}^{\mathcal{G}_c}T(\Psi) {}^{\mathcal{G}_p}M$, 
where ${}_{\mathcal{G}_p}^{\mathcal{G}_c}T(\Psi)$ is a transformation matrix representing the camera’s rigid motion.
In the current frame, this point is projected onto a point 
$m_c = \pi({}^{\mathcal{G}_c}M; {}_{\mathcal{C}_c}^{\mathcal{G}_c}\Theta_c)$
on the image plane.
The POC estimates the difference between $m_p$ and $m_c$, and we minimize the difference between $d$ and $ m_p - m_c $.

We solved the minimization problem as follows:
\begin{align}
    \{\Theta, \Psi\} = \argmin_{\Theta, \Psi} \| m_c - m_p - d \|^2. \label{eq:optimOnImage}
\end{align}
We minimized the objective function by the Levenberg--Marquardt method.
An example of the function in postural and translational slices is illustrated in Fig.~\ref{fig:process}, which shows the optimization is insensitive to the initial values; most can be set to zero.

In the minimization process, obvious outliers were rejected based on the magnitude of the estimated displacement vector.
For greater robustness against outliers, we selected a predetermined number (50 in the experiment) of subsets randomly sampled from the remaining vectors after the outlier rejection in a given ratio (0.6 in the experiments). The best parameters that gave the smallest residuals were picked as the estimates.

\subsectionskip
\subsection{Refinement via virtual IPM}
The initial estimates were refined via virtual IPM. We mapped the interest points on the image to the virtual IPM plane based on the current pose estimation as shown in Fig.~\ref{fig:workflow}. In addition, we cropped a set of patches centered at these mapped points from the inversely mapped captured image (virtual IPM image). The size of the patches was predetermined; however, some fixed-size patches possibly protruded into the boundary in the marginal area of the virtual IPM image. We shrank those patches such that the whole region fell within the virtual IPM image as exemplified in Fig.~\ref{fig:workflow}. After we obtained a motion vector field on the virtual IPM plane, we minimized the reprojection residuals on the virtual IPM plane. 

Because the computed vector field was on the virtual IPM, not the IPM image based on the true pose, we explicitly distinguished the estimated pose from the true pose.
Virtual IPM images $\hat{\mathcal{I}}_p$ and $\hat{\mathcal{I}}_c$ were computed by current pose estimates $\hat{\Theta}_p$ and $\hat{\Theta}_c$ with respect to ground coordinate systems $\hat{\mathcal{G}}_p$ and $\hat{\mathcal{G}}_c$, respectively (Fig.~\ref{fig:parameterization}).
First, we selected point ${}^{\hat{\mathcal{G}}_p}M_p$ on virtual IPM image $\hat{\mathcal{I}}_p$, which was inversely projected from an interest point. This point was associated with point ${}^{\mathcal{G}_p} M_p$ on the ground with respect to estimated and true pose parameters as follows:
\begin{align}
    {}^{\mathcal{G}_p} M_p = \pi^{-1}(\pi({}^{\hat{\mathcal{G}}_p} M_p;~{}_{C_p}^{\hat{\mathcal{G}}_p}\hat{\Theta}_p);~{}_{C_p}^{\mathcal{G}_p} \Theta_p).
\end{align}
This point should be observed in the current frame's virtual IPM plane as
\begin{align}
    {}^{\hat{\mathcal{G}}_c}M_c = \pi^{-1}(
        \pi({}_{\mathcal{G}_p}^{\mathcal{G}_c}T(\Psi) {}^{\mathcal{G}_p}M_p;~
        {}_{C_c}^{\mathcal{G}_c}\Theta_c);~
    {}_{C_c}^{\hat{\mathcal{G}}_c}\hat{\Theta}_c)),
\end{align}
which is a point that POC estimates as a displaced location of ${}^{\hat{\mathcal{G}}_p} M_p$.
Because we had an estimated motion vector field on the virtual IPM planes, we minimized the squared difference between the motion vector field and estimated displacement, as described earlier.
We formulated the minimization problem as follows:
\begin{align}
    \{\Theta, \Psi\} = \argmin_{\Theta, \Psi} \| {}^{\hat{\mathcal{G}}_c}M_c - {}^{\hat{\mathcal{G}}_p}M_p - \hat{d} \|^2,
\end{align}
where $\hat{d}$ is the estimated displacement corresponding to ${}^{\hat{\mathcal{G}}_p}M_p$.
This minimization problem can be solved stably as suggested by the shape of the objective function (Fig.~\ref{fig:process}).
Outliers were rejected based on displacement vectors induced by previously estimated parameters.

\begin{figure*}[t]
    \centering
    \includegraphics[width=\linewidth]{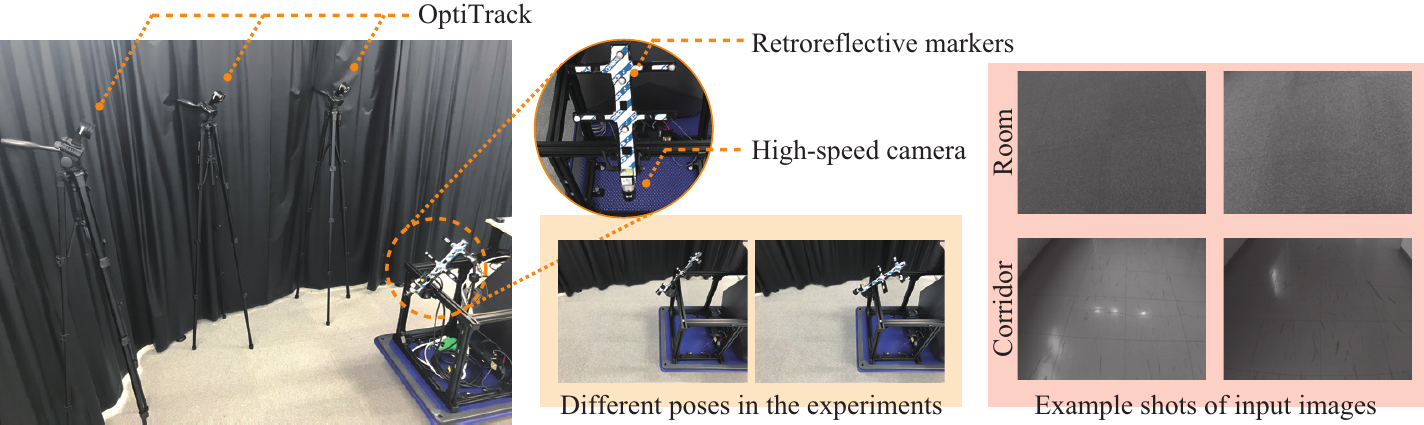}
    \figureskip
    \caption{Experimental setup. A camera is solidly mounted on a hand cart with retroreflective markers for ground truth evaluation. The insets show example shots of different poses and captured images.}
    \label{fig:system}
    \figureskip
\end{figure*} 

\sectionskip
\section{Evaluation}
\sectionskip
\subsection{Experimental setup}
In Fig.~\ref{fig:system}, the experimental system is shown. The camera was mounted on a solid aluminum frame immobilized on a hand cart at a height of approximately 700 mm. We used an industrial camera, acA800-510um, manufactured by Basler Inc. (Resolution: $800 \times 600$ px) with a lens (F1.4 / 3.5 mm.) We set the frame rate to 100 fps and downsampled to 10 fps. For ground-truth evaluation, we employed six 3D-motion-capture cameras (OptiTrack Flex13 by NaturalPoint Inc.\cite{OptiTrack}) with a carefully designed retroreflective marker plate attached to the camera, as shown in Fig.~\ref{fig:system}. The mean 3D localization error was approximately 0.2 mm.
In the experiments, we set the patch size as 128 and 256 on the image and virtual IPM planes, respectively.
We used 11 by 9 interest points for the experiments.
In the initial optimization, we set the initial pitch angle to $\pi/3$ and set the other parameters at 0 for all experiments.
Fig.~\ref{fig:system} shows an example sequence of the captured images.

For quantitative comparison, a state-of-the-art visual odometry method, semi-direct visual odometry (SVO2) \cite{Forster17}, was used. 
Because optimal performance is achieved with a high-speed camera (more than 70 fps as authors suggested),
captured image sequences without downsampling (100 fps) were used.
Note that we confirmed it failed for all image sequences at 10 fps.
Algorithm parameters were set to their defaults.
Thus, comparison was performed after SVO2 successfully completed the initialization process.
Pitch and roll angles at the first frame and the scaling parameter were determined by the motion-capture system.

\begin{table}[b]
\tableskip
\centering
\caption{Estimation errors of pitch [deg], roll [deg], and travel distance (Dist.) [mm] in initial and refinement steps. The mean and standard deviation are shown as {\it Mean(Std dev).} (Rm: Room, Cor: Corridor)}
\label{table:pitchRollTDError}
\small
\begin{tabular}{|c|c|c|c|c|}
\hline
\multicolumn{2}{|c|}{}                    & \multicolumn{3}{c|}{Iteration}    \\ \cline{3-5} 
\multicolumn{2}{|c|}{}                    & Initial & 1st refine & 2nd refine \\ \hline
\multirow{3}{*}{Rm.} & Pitch   & 0.538(0.355) & 0.421(0.278)  & 0.346(0.338) \\       
                      & Roll   & 0.887(0.455) & 0.642(0.430)  & 0.609(0.484) \\       
                      & Dist.  & 1.03(0.344)  & 0.234(0.171)  & 0.171(0.118) \\ \hline
\multirow{2}{*}{Cor.} & Pitch   & 1.88(1.43)   & 0.806(0.716)  & 0.857(0.558) \\
                      & Roll    & 2.98(2.55)   & 1.21(0.819)   & 1.23(0.884)  \\ \hline
\end{tabular}
\end{table}

\begin{figure*}
    \centering
    \includegraphics[width=\linewidth]{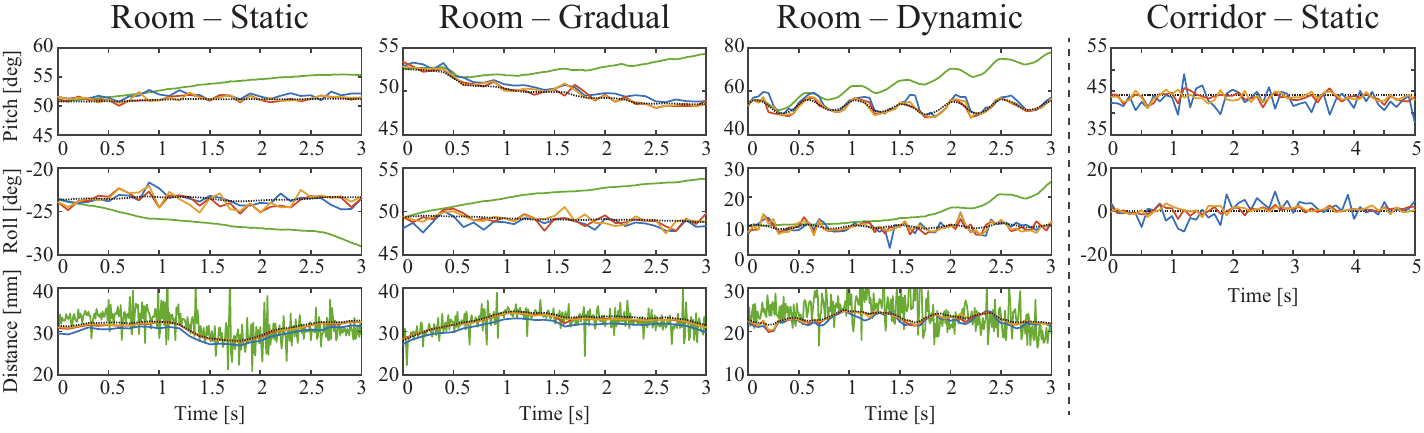}
    \caption{Estimation results. The cameras moved in a static pose and two dynamically changing poses in the room and corridor. The blue, red, and yellow lines show the initial estimation, estimations after one refinement, and estimations after two refinements, respectively. The green lines show the SVO2 estimation result. The figure shows that the refinement process improves estimation accuracy. The black dotted line shows the ground truth.}
    \label{fig:pitchRollTD}
    \figureskip
\end{figure*} 

\subsectionskip
\subsection{Experimental results}
We moved the hand cart for three different time-varying poses: a static pose, a gradually changing pose, and a dynamically changing pose in a room.
We manually tilted the camera during the motion to change the pose by grabbing the marker plate.
Fig.~\ref{fig:pitchRollTD} (left three) shows the estimation results of pitch and roll angles $\Theta$ and travel distance $\sqrt{tx^2 + tz^2}$.
These figures show that the proposed method successfully estimated the pose and motion with an accuracy of approximately $0.5$ degrees for pitch and roll angles and an accuracy of approximately $0.2$ mm for the travel distance in a static pose and a gradually changing pose. Even when the camera was shaken for a short period, as demonstrated on the right in Fig.~\ref{fig:pitchRollTD}, the proposed method could estimate pose and motion, in which the mean error of the pitch and roll angles was below $1.0$ degrees and that of the travel distance was below $0.3$ mm. 
Thus, the assumption holds that a camera's vertical motion within a frame interval is ignorable compared to camera's height.
It also shows the estimation accuracy was improved through the refinement process. In the conducted experiments, we empirically confirmed one refinement is practically sufficient to obtain good results. 
Statistics of the estimation in the experiments demonstrate that the refinement process with virtual IPM reduced the estimation error (Table~\ref{table:pitchRollTDError}).
However, SVO2 experienced cumulative errors in pitch and roll angle estimation and large noise in travel distance estimation. These deteriorations were mainly due to a lack of a reference structure and inaccurate displacement estimation of feature points compared to the proposed method.

\begin{figure}[t]
    \centering
    \includegraphics[width=\linewidth, keepaspectratio]{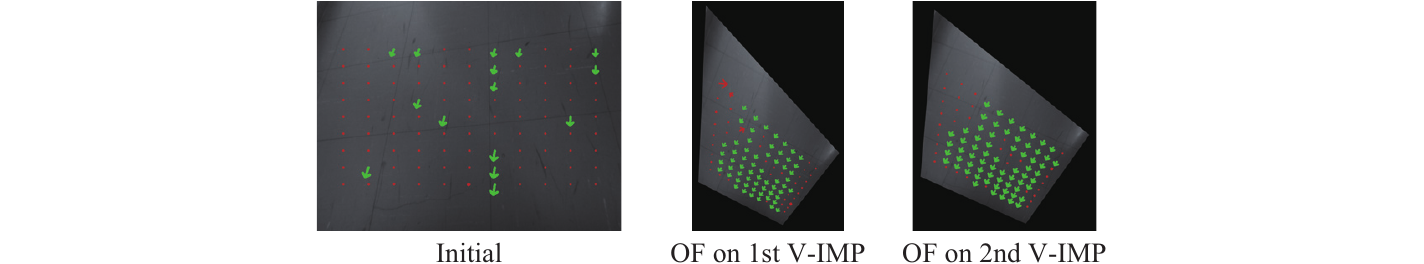}
    \caption{Example procedure. Inliers and outliers are colored in green and red, respectively.}
    \label{fig:processExample}
    \figureskip
\end{figure}

Fig.~\ref{fig:pitchRollTD} (rightmost) shows the estimation results when the camera was statically moved in a corridor. We observed highlights and reflections on the floor under dynamically changing lighting conditions as shown in Fig.~\ref{fig:system}.
Statistical results shown in Table~\ref{table:pitchRollTDError} also show the proposed method greatly refines estimated pitch and roll angles. 
This shows the proposed framework using virtual IPM works well even in the case of a very weak ground texture.
Fig.~\ref{fig:processExample} demonstrates optical flows on image and virtual IPMs. Successful POC estimation on the virtual IPM increased in number compared to that on the image, resulting in much better accuracy.
In contrast, SVO2 failed in estimation. This is chiefly because extracted feature points could not track the same point owing to strong highlights and weak a texture in sight, which resulted in failure at converging parameters. 
Note that we used constant pitch and roll angles estimated by camera calibration as ground truth in the corridor experiment.

\subsectionskip
\subsection{Discussion}
The proposed method achieved this level of accuracy with only two frames. This property is particularly valuable when a camera pose dynamically changes while in motion, as exemplified in Fig.~\ref{fig:pitchRollTD}. 
Moreover, our method can be further extended for use in multiple frames and Bayes filtering techniques to gain more robustness and accuracy.
Currently, the method requires time owing to a prototyping implementation; we expect high-speed processing according to the parallelizable architecture.

\sectionskip
\section{Conclusion}
\sectionskip
We proposed an autocalibration and motion estimation method for a monocular camera mounted on a ground-moving robot.
The method adopts a semi-dense approach, which favorably combines a feature-based method and an image-registration technique, taking advantage of a ground-moving robot-mounted camera configuration.
To make maximum use of the highly accurate and efficient image-registration techniques proposed for 2D rigid-motion estimation, we introduced refinement steps by setting virtual IPM planes to eliminate the perspective effect.
We achieved accurate estimation, in which the relative mean error of the pitch and roll angles was approximately $1.0$ degrees and the absolute mean error of travel distance was $0.3$ mm, even in a highly dynamic scene.

Currently, the proposed system deteriorates in performance if there is an obstacle on the nearby ground.
We will integrate obstacle detection modules to overcome this limitation.
In the future, we also plan to improve the proposed method with respect to parallelization, multi-camera use, Bayes filtering, and nonholonomic constraints.
We look forward to implementing the proposed method for an automobile to enhance advanced driver-assistance systems and autonomous driving.




\bibliographystyle{IEEEtran}
\bibliography{egbib}

\begin{thebibliography}{10}
\providecommand{\url}[1]{#1}
\csname url@rmstyle\endcsname
\providecommand{\newblock}{\relax}
\providecommand{\bibinfo}[2]{#2}
\providecommand\BIBentrySTDinterwordspacing{\spaceskip=0pt\relax}
\providecommand\BIBentryALTinterwordstretchfactor{4}
\providecommand\BIBentryALTinterwordspacing{\spaceskip=\fontdimen2\font plus
\BIBentryALTinterwordstretchfactor\fontdimen3\font minus
  \fontdimen4\font\relax}
\providecommand\BIBforeignlanguage[2]{{%
\expandafter\ifx\csname l@#1\endcsname\relax
\typeout{** WARNING: IEEEtran.bst: No hyphenation pattern has been}%
\typeout{** loaded for the language `#1'. Using the pattern for}%
\typeout{** the default language instead.}%
\else
\language=\csname l@#1\endcsname
\fi
#2}}

\bibitem{Lee16}
T.~Lee, D.~Yi, and D.~Cho, ``A monocular vision sensor-based obstacle detection
  algorithm for autonomous robots,'' \emph{Sensors}, vol.~16, no.~3, 2016.

\bibitem{Bertozzi98a}
M.~Bertozzi and A.~Broggi, ``{GOLD}: a parallel real-time stereo vision system
  for generic obstacle and lane detection,'' \emph{IEEE Transactions on Image
  Processing}, vol.~7, no.~1, pp. 62--81, 1998.

\bibitem{Lovegrove11}
S.~Lovegrove, A.~J. Davison, and J.~I. {n}ez Guzm\'{a}n, ``Accurate visual
  odometry from a rear parking camera,'' \emph{IEEE Intelligent Vehicles
  Symposium}, pp. 788--793, 2011.

\bibitem{Westerhoff16}
J.~{Westerhoff}, S.~{Lessmann}, M.~{Meuter}, J.~{Siegemund}, and A.~{Kummert},
  ``Development and comparison of homography based estimation techniques for
  camera to road surface orientation,'' in \emph{IEEE Intelligent Vehicles
  Symposium}, 2016, pp. 1034--1040.

\bibitem{Nister04}
D.~{Nister}, O.~{Naroditsky}, and J.~{Bergen}, ``Visual odometry,'' in
  \emph{IEEE International Conference on Computer Vision and Pattern
  Recognition}, vol.~1, 2004.

\bibitem{Forster17}
C.~{Forster}, Z.~{Zhang}, M.~{Gassner}, M.~{Werlberger}, and D.~{Scaramuzza},
  ``{SVO}: Semidirect visual odometry for monocular and multicamera systems,''
  \emph{IEEE Transactions on Robotics}, vol.~33, no.~2, pp. 249--265, 2017,
  http://rpg.ifi.uzh.ch/svo2.html (accessed on April 22nd, 2020).

\bibitem{Chen94}
Q.~Chen, M.~Defrise, and F.~Deconinck, ``Symmetric phase-only matched filtering
  of {Fourier-Mellin} transforms for image registration and recognition,''
  \emph{IEEE Transactions on Pattern Analysis and Machine Intelligence},
  vol.~16, no.~12, pp. 1156--1168, 1994.

\bibitem{Takita03}
K.~Takita, T.~Aoki, Y.~Sasaki, T.~Higuchi, and K.~Kobayashi, ``High-accuracy
  subpixel image registration based on phase-only correlation,'' \emph{IEICE
  Transactions on Fundamentals of Electronics, Communications and Computer
  Sciences}, vol.~86, no.~8, pp. 1925--1934, 2003.

\bibitem{Bertozzi98b}
M.~Bertozzi, A.~Broggi, and A.~Fascioli, ``Stereo inverse perspective mapping:
  theory and applications,'' \emph{Image and Vision Computing}, vol.~16, no.~8,
  pp. 585 -- 590, 1998.

\bibitem{Miksch10}
M.~Miksch, B.~Yang, and K.~Zimmermann, ``Automatic extrinsic camera
  self-calibration based on homography and epipolar geometry,'' \emph{IEEE
  Intelligent Vehicles Symposium}, pp. 832--839, 2010.

\bibitem{Knorr13}
M.~Knorr, W.~Niehsen, and C.~Stiller, ``Online extrinsic multi-camera
  calibration using ground plane induced homographies,'' \emph{IEEE Intelligent
  Vehicles Symposium}, pp. 236--241, 2013.

\bibitem{Zienkiewicz15}
J.~Zienkiewicz and A.~Davison, ``Extrinsics autocalibration for dense planar
  visual odometry,'' \emph{Journal of Field Robotics}, vol.~32, no.~5, pp.
  803--825, 2015.

\bibitem{Campbell05}
J.~{Campbell}, R.~{Sukthankar}, I.~{Nourbakhsh}, and A.~{Pahwa}, ``A robust
  visual odometry and precipice detection system using consumer-grade monocular
  vision,'' in \emph{IEEE International Conference on Robotics and Automation},
  2005, pp. 3421--3427.

\bibitem{NouraniVatani09}
N.~Nourani-Vatani, J.~M. Roberts, and M.~V. Srinivasan, ``Practical visual
  odometry for car-like vehicles,'' \emph{IEEE International Conference on
  Robotics and Automation}, pp. 3551--3557, 2009.

\bibitem{Kitt11}
B.~Kitt, J.~Rehder, A.~Chambers, M.~Sch{\"o}nbein, H.~Lategahn, and S.~Singh,
  ``Monocular visual odometry using a planar road model to solve scale
  ambiguity,'' in \emph{European Conference on Mobile Robotics}, 2011.

\bibitem{Malis04}
E.~{Malis}, ``Improving vision-based control using efficient second-order
  minimization techniques,'' in \emph{IEEE International Conference on Robotics
  and Automation}, vol.~2, 2004, pp. 1843--1848.

\bibitem{OptiTrack}
{NaturalPoint Inc.}, ``{OptiTrack Flex13},'' https://www.optitrack.com/
  (accessed on January 30th, 2020).

\end{thebibliography}

\end{document}